# JOINT RNN-BASED GREEDY PARSING AND WORD COMPOSITION

**Joël Legrand**
Idiap Research Institute, Martigny, Switzerland
École Polytechnique Fédérale de Lausanne (EPFL), Lausanne Switzerland
`joel.legrand@idiap.ch`

**Ronan Collobert** *
Facebook AI Research, Menlo Park, CA, USA
Idiap Research Institute, Martigny, Switzerland
`ronan@collobert.com`

## ABSTRACT

This paper introduces a greedy parser based on neural networks, which leverages a new compositional sub-tree representation. The greedy parser and the compositional procedure are jointly trained, and tightly depends on each-other. The composition procedure outputs a vector representation which summarizes syntactically (parsing tags) and semantically (words) sub-trees. Composition and tagging is achieved over continuous (word or tag) representations, and recurrent neural networks. We reach F1 performance on par with well-known existing parsers, while having the advantage of speed, thanks to the greedy nature of the parser. We provide a fully functional implementation of the method described in this paper [1].

## 1 INTRODUCTION

In Natural Language Processing (NLP), the parsing task aims at analysing the underlying syntactic structure of a natural language sequence of words (a sentence). The analysis is expressed as a tree of syntactic relations between sub-constituents of the sentence. In the linguistic world, Chomsky (1956) first introduced formally the parsing task, by defining the natural language syntax as a set of context-free grammar rules (a particular type of formal grammar), combined with transformations rules. Automated syntactic parsing became rapidly a key task in computational linguistic. A parse tree not only carries syntax information, but might also embed some semantic information (in the sense that it can disambiguate different interpretations of a given sentence). In that respect, parsing it has been widely used as an input feature for several other NLP tasks such as machine translation (Zollmann & Venugopal, 2006), information retrieval (Alonso et al., 2002), or Semantic Role Labeling (Punyakanok et al., 2008).

This paper introduces a greedy parser which leverages a new composition approach to keep an history of what has been predicted so far. The composition performs a syntactic and semantic summary of the contents of a sub-tree in the form of a vector representation. The composition is performed along the tree: bottom tree node representations are obtained by composing *continuous word vector representations*, and produces vector representations which are in turn composed together in subsequent nodes of the tree. The composition operation as well as tree node tagging and predictions are achieved with a Recurrent Neural Network (RNN). Both the composition and node prediction are trained *jointly*.

Section 2 presents several related approaches. Section 3 details our parsing architecture. An empirical evaluation of our models as well as our compositional vectors is given in Section 4.

---

*All research was conducted at the Idiap Research Institute, before Ronan Collobert joined Facebook AI Research.
[1]The parser can be downloaded at `joel-legrand.fr/parser`.

## 2 RELATED WORK

The first attempts to automatically parse natural language were mainly conducted using generative models. A wide range of parser were, and still are, based on Probabilistic context-free grammar (PCFGs) (Magerman, 1995; Collins, 2003; Charniak, 2000). These types of parsers model the syntactic grammar by computing statistics of simple grammar rules (over parsing tags) occurring in a training corpus. However, many language ambiguities cannot be caught with simple tag-based PCFG rules. A key element in the success of PCFGs is to refine the rules with a word lexicon. This is usually achieved by attaching to PCFGs a lexical information called the *head-word*. Several head-word variants exist, but they all rely on a deterministic procedure which leverages clever linguistic knowledge. Parsing inference is mostly achieved using simple bottom-up chart parser (Kasami, 1965; Earley, 1970; Kay, 1986). These methods face a classical learning dilemma: on one hand PCFG rules have to be refined enough to avoid any ambiguities in the prediction. On the other hand, too much refinement in these rules implies lower occurrences in the training set, and thus a possible generalization issue. PCFGs-based parsers are thus judiciously composed with carefully chosen PCFG rules and clever regularization tricks.

### 2.1 STATE-OF-THE-ART

Discriminative approaches from Henderson (2004); Charniak & Johnson (2005) outperform standard PCFG-based generative parsers, but only by discriminatively *re-ranking* the $K$-best predicted trees coming out of a generative parser. To our knowledge, the state of the art in syntactic parsing is still held by McClosky et al. (2006), who leverages discriminative re-ranking, as well as self-training over unlabeled corpora: a re-ranker is trained over a generative model which is then used to label the unlabeled dataset. The original parser is then re-trained with this new "labeled" corpus. Petrov & Klein (2007) introduced a method to automatically refine PCFG rules by iteratively splitting them. This method leverages an efficient coarse-to-fine procedure to speed up the decoding process. More recently, Finkel et al. (2008); Petrov & Klein (2008) proposed PCFG-based *discriminative* parsers reaching the performance of their generative counterparts. Conditional Random Fields (CRFs) are at the core of such approaches. Carreras et al. (2008) currently holds the state-of-the-art among the (non-reranking) discriminative parsers. Their parser leverages a global-linear model (instead of a CRF) with PCFGs, together with various new advanced features. Z. et al. (2010) showed that jointly using multiple self-trained grammars can achieve higher accuracy than an individual grammar.

In contrast to these existing approaches, our parser does not rely on PCFGs, nor on refined features like head-words. Tagging nodes is achieved in a greedy manner, using only raw words and part-of-speech (POS) as features. Tree node history is maintained as a vector representation obtained in a recurrent fashion, by composing past node representations and tag predictions.

### 2.2 GREEDY PARSING

Many discriminative parsers follows a greedy strategy because of the lack (or the intractability) of a global tree score for an entire derivation path which would combine independent node decisions. Adopting a greedy strategy that maximize local scores for individual decisions is then a solution worth investigating. One of the first successful discriminative parsers (Ratnaparkhi, 1999) was based on MaxEnt classifiers (trained over a large number of different features) and powered a greedy shift-reduce strategy.

Henderson (2003) introduced a generative left-corner parser where the probability of a derivation given the derivation historic was approximated using a Simple Synchrony Networks (SNN), which is a neural network specifically designed for processing structures.

Turian & Melamed (2006) later proposed a bottom-up greedy algorithm following a left-to-right or a right-to left strategy and using using a feature boosting approach. In this approach, greedy decisions regarding the tree construction are made using decision tree classifiers. Their model was nevertheless limited to short length sentences.

Zhu et al. (2013) proposed a shift-reduce parser which achieves results comparable to their chart-based counterparts. This is done by leveraging several unsupervisely trained features (word Brown clustering, dependency relations, dependency language model) combined with a smart beam search strategy.

### 2.3 PARSING WITH RECURRENT NEURAL NETWORKS

Recurrent Neural Networks (RNNs) were seen very early (Elman, 1991) as a way to tackle the problem of parsing, as they can naturally recur along the parse tree. A first practical application of RNN on syntactic parsing were proposed by Costa et al. (2002). Their approach was based on a left-to-right incremental parser, where a recursive neural network was used to re-rank possible phrase attachments. The goal of their contribution was, in their own terms, the assessment of a methodology rather than a fully functional system. They demonstrated that RNNs were able to capture enough information to make correct parsing decisions.

Collobert (2011) proposed a purely discriminative parser based on neural networks. This model leveraged continuous vector representations from Collobert & Weston (2008), and builds the full parsing tree in a bottom-up manner. To deal with the recursive structure inherent to syntactic parsing, a very simple history was given to the network as a new vector feature (corresponding to the nearest tag spanning the word being tagged).

Socher et al. (2011) also leveraged continuous vectors from Collobert & Weston (2008), combining them to build a tree in a greedy manner. However, this work did not tackle the full parse tree problem, but was restricted to unlabeled bracketing. Socher et al. (2013) introduced the Compositional Vector Grammar (CVG) which combines PCFGs with a Syntactically Untied Recursive Neural Network (SU-RNN). Composition is performed over a binary tree, then used to score the $K$-best trees coming out of a generative parser. For a given (parent) node of the tree, the authors apply a composition operation over its child nodes, conditioned with their syntax information. In contrast, we compose phrases (not limited to two words). Both the words and syntax information of the child nodes are fed to each composition operation, leading to a vector representation of each tree node carrying both some semantic and syntactic information. We also do not rely on any generative parser as our model *jointly* trains the task of node prediction, and the task of node composition.

Legrand & Collobert (2014) proposed a greedy RNN-based parser. The neural network was recurrent only in the sense it used previously predicted tags to produce next tree node tags. Contrary to Socher et al. (2013), it did not involve composing sub-tree representations. Instead, head-words were used as a key feature. Our approach shares some similarities with (Legrand & Collobert, 2014), as it is also a greedy parser based on RNNs. However, instead of relying on head-words (which could be seen as a simplistic representations of sub-trees), we leverage compositional sub-tree vector representations trained jointly with the parser. This approach leads to much better parsing accuracy, while relying only on a few simple features (words and POS tags). Our model has also the ability of producing phrase embeddings, which may represent a valuable feature for other NLP tasks.

Chen & Manning (2014) proposed a greedy transition-based dependency parser based on neural networks, fed with dense word and tag vector representations. In contrast to our approach, it does not integrate a compositional procedure over sentence sub-trees. The network is only involved in predicting correct transitions at each step of the parsing process.

## 3 GREEDY RNN PARSING

Our parser is based on a neural network tagger, and perform parsing in a greedy recurrent way. Our approach is a bottom-up iterative procedure: the tree is constructed starting from the terminal nodes (sentence words), as shown in Figure 1. At each iteration,

1. We look for all possible new tree nodes merging input constituents (i.e., heads of the trees predicted so far or leaves which have not been composed so far). For that purpose, we apply a neural network (see Figure 3) sliding window tagger over input constituents $X_1, \ldots, X_N$. Considering an arbitrary rule

$$A \to X_i, X_{i+1}, \ldots, X_j$$

| | | | | | | | | |
|---|---|---|---|---|---|---|---|---|
| $\mathbf{I}_W$ : | Look | around | and | choose | your | own | ground | . |
| $\mathbf{I}_T$ : | VB | RP | CC | VB | PRP$ | JJ | NN | . |
| **O** : | O | S-PRT | O | O | B-NP | I-NP | E-NP | O |
| $\mathbf{I}_W$ : | Look | *R1* | and | choose | *R2* | . | | |
| $\mathbf{I}_T$ : | VB | PRT | CC | VB | NP | . | | |
| **O** : | B-VP | E-VP | O | B-VP | E-VP | O | | |
| $\mathbf{I}_W$ : | *R3* | and | *R4* | . | | | | |
| $\mathbf{I}_T$ : | VP | CC | VP | . | | | | |
| **O** : | B-VP | I-VP | E-VP | O | | | | |
| $\mathbf{I}_W$ : | *R5* | . | | | | | | |
| $\mathbf{I}_T$ : | VP | . | | | | | | |
| **O** : | B-S | E-S | | | | | | |

Figure 1: Greedy parsing algorithm, on the sentence "Look around and choose your own ground.". $\mathbf{I}_W$, $\mathbf{I}_T$ and **O** stand for input words (or composed word representations $R_i$), input syntactic tags (parsing or part-of-speech) and output tags (parsing), respectively. See Figure 2 and Section 3.2 for the word composition procedure. The tree produced after 4 greedy iterations (as shown here) can be reconstructed as the following: (S (VP (VP (VB Look) (PRT (RP around))) (CC and) (VP (VB choose) (NP (PRP$ your) (JJ own) (NN ground)))) (. .)).

   defining a new node with tag $A$, the tagger will produce prefixed tags $B$-$A$, $I$-$A$, ... $E$-$A$, respectively for constituents $X_i$, $X_{i+1}$, ..., $X_j$, following a classical BIOES prefixing scheme[2].

2. A simple dynamic programming is performed, only to insure the coherence of the tag prediction (e.g., a $B$-$A$ can be followed only by a $I$-$A$ or a $E$-$A$).

3. A (neural network) composition module computes vector representations of the new nodes, according to the representations of the merged constituents, as well as the tag predictions (see Figure 2).

4. New predicted nodes become input constituents and we go back to 1 (see Figure 1).

Our system is recurrent in two ways: newly predicted parsing node labels as well as vector representations obtained by composing these predicted nodes, are used in the next iteration of our algorithm.

We will detail our architecture in the following.

### 3.1 WORD EMBEDDINGS

Following the work from Collobert & Weston (2008) on various NLP tasks, our parser relies on *raw words*. Each word in a finite dictionary $\mathcal{W}$, is assigned a continuous vector representation. These representations as all parameters of our architecture are trained by back-propagation. More formally, given a sentence of $N$ words, $w_1, w_2, ..., w_N$, each word $w_n \in \mathcal{W}$ is first embedded in a $D$-dimensional vector space by applying a lookup-table operation:

$$LT_W(w_n) = W_{w_n},$$

where the matrix $W \in \mathbb{R}^{D \times |\mathcal{W}|}$ represents the parameters to be trained in this lookup layer. Each column $W_n \in \mathbb{R}^D$ corresponds to the vector embedding of the $n^{th}$ word in our dictionary $\mathcal{W}$.

In this work, two kind of features are used to feed the networks: words (or word compositions) and POS tags $\mathcal{T}$ (or parsing tags $\mathcal{P}$). As for words, a lookup-table associates each tag $t$ in the finite set of tag $\mathcal{T} \cup \mathcal{P}$ with a continuous vector representation of size $T$. The output vectors of the different lookup-tables are simply concatenated to form the input of the next layer.

---
[2]B*egin*, I*ntermediate*, O*ther*, E*nd*, S*ingle*. This approach is very often used in NLP, when one wants to rewrite a chunk (here node) prediction problem into a word tagging problem.

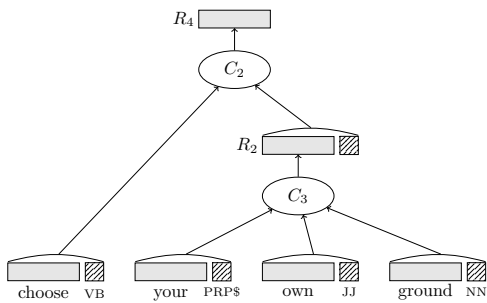
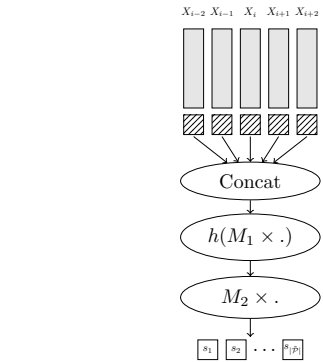

Figure 2: Recurrent composition of the sub-tree (VP (VB choose) (NP (PRP$ your) (JJ own) (NN ground))).
The representation R2 is first computed using the 3-inputs module $C_3$ with your/PRP$ own/JJ ground/NN as input. R4 is obtained by using the 2-inputs module $C_2$ with choose/VB R1/NP as input

Figure 3: A constituent $X_i$ is tagged by considering a fixed size context window of size $K$ (here $K = 5$). The concatenated output of the compositional history and constituent tags is fed as input to the tagger. It outputs a score for each BIOES-prefixed parsing tag. The tagger is a standard two-layer neural network. Tags for the current sequence of constituents $X_1, \ldots, X_N$ is obtained by simply sliding this network over the sequence.

Using continuous word vectors as input allows us to take advantage of unsupervisely pre-trained word embeddings. Lot of work on this domain has been done in recent year, including Collobert & Weston (2008), Mikolov et al. (2013). In this paper, we chose to use the representations from Lebret & Collobert (2014), obtained by a simple PCA on a matrix of word co-occurrences.

### 3.2 WORD-TAG COMPOSITION

At each step of the parsing procedure, we represents each node of the tree as a vector representation, which summarizes both the syntax (predicted POS or parsing tags) and the semantic (words) of the sub-tree corresponding to the given node. As shown in Figure 2, the vector representation is obtained by a simple recurrent procedure, which involves several components:

- Word vector representations *for the leaves* (coming out from a lookup table) (dimension $D$).
- Tag (POS for the leaves, predicted tags otherwise) vector representations (also coming out for another lookup table, as explained in Section 3.1) (dimension $T$).
- Compositional networks $C_k()$. Each of them can compress the representation of a chunk of size $k$ into a $D$-dimensional vector.

Compositional networks take as input both the merged node representations and predicted tag representations. There is one different network $C_k$ for each possible node with a number of $k$ merged constituent. In practice most tree nodes do not merge more than a few constituents[3]. In our case, denoting $z \in \mathbb{R}^{(D+T) \times k}$ the concatenation of the merged constituent representations ($k$ vectors of tags and constituent representations), the compositional network is simply a matrix-vector operation followed by a non-linearity

$$C_k(z) = h(M^k z),$$

where $M^k \in \mathbb{R}^{D \times (k(D+T))}$ is a matrix of parameters to be trained, and $h()$ is a simple non-linearity such as a pointwise hyperbolic tangent.

---
[3] Taking $1 \leq k \leq 5$ covers already 98.6% of the nodes in the Wall Street Journal training corpus, and $1 \leq k \leq 7$ covers 99.8%.

Note that node and word representations are embedded in the same space. This way, the compositional networks $C_k$ can compress indifferently information coming from leaves or sub-trees. Implementation-wise, one can store new node representations into the word lookup-table as the tree is created, such that subsequent composition or tagging operations can be achieved in an efficient manner.

### 3.3 SLIDING WINDOW BIOES TAGGER

The tagging module of our architecture (see Figure 3) is a two-layer neural network which applies a sliding window over the input constituent representations (as computed in Section 3.2), as well as the input constituent tag representations. Considering $N$ input constituents $X_1, \ldots, X_N$, if we assume their respective representations has been stored so far in lookup tables, the $n^{th}$ window is defined as
$$u_n = [LT(X_{n-\frac{K-1}{2}}), ..., LT(X_n), ..., LT(X_{n+\frac{K-1}{2}})],$$
where $K$ is the size of window. Denoting $\tilde{\mathcal{P}}$ the set of BIOES-prefixed parsing tags from $\mathcal{P}$, the module outputs a vector of scores $s(u_n) = [s_1, ..., s_{|\tilde{\mathcal{P}}|}]$ (where $s_t$ is the score of the BIOES-prefixed parsing tag $t \in \tilde{\mathcal{P}}$ for the constituent $X_n$). The constituent with indices exceeding the input boundaries ($n - (K-1)/2 < 1$ or $n + (K-1)/2 > N$) are mapped to a special padding vector (which is also learned). As any classical two-layer neural network, our architecture performs several matrix-vector operations on its inputs, interleaved with some non-linear transfer function $h(\cdot)$,
$$s(u_n) = M_2 \, h(M_1 \, u_n),$$
where the matrices $M_1 \in \mathbb{R}^{H \times K|D|}$ and $M_1 \in \mathbb{R}^{|\tilde{\mathcal{P}}| \times H}$ are the trained parameters of the network, and $h()$ is a pointwise non-linear function such as the hyperbolic tangent. The number of hidden units H is a hyper-parameter to be tuned.

### 3.4 COHERENT BIOES PREDICTIONS

The next module of our architecture aggregates the BIOES-prefixed parsing tags from our tagger module in a coherent manner. It is implemented as a Viterbi decoding algorithm over a constrained graph $G$, which encodes all the possible valid sequences of BIOES-prefixed tags over constituents: e.g. $B$-$A$ tags can only be followed by $I$-$A$ or $E$-$A$ tags, for any parsing label $A$. Each node of the graph is assigned a score produced by the previous neural network module (score for each BIOES-prefixed tag, and for each word). The score $S([t]_1^N, [X]_1^N, \theta)$ for a sequence of tags $[t]_1^N$ in the lattice $G$ is simply obtained by summing scores along the path ($[X]_1^N$ being the input sequence of constituents and $\theta$ all the parameters of the model). We followed the exact same approach as in (Legrand & Collobert, 2014), except that transition scores (edges on the graph) were all set to zero. Indeed, we observed in empirical experiments that adding transitions scores does not improve F1-score performance. This decoding is thus present only to insure coherence in the predicted sequence of tags.

### 3.5 TRAINING PROCEDURE

Both the composition network and tagging networks are trained by maximizing a likelihood over the training data using stochastic gradient ascent.

We performed all possible iterations, over all training sentences, of the greedy procedure presented in Figure 1 constrained with the provided labeled parse tree. This leads to our training set of sequences of tree nodes. While this procedure is similar to (Legrand & Collobert, 2014), it is worth mentioning that it implies the system is only trained on *correct* sequences of tree nodes. In that respect, it is not trained to recover from past mistakes it could have made during the recurrent process. For every tree node, the sub-trees (structure and tags) were also extracted during this procedure.

Training the system consists in repeating the following steps:

- Pick a random sequence of nodes extracted in the training set, as described above. Consider the associated sub-trees for each node which is not a leaf.
- Perform a forward pass of the word-tag composer (see Section 3.2) along these sub-trees.

- For all nodes in the sequence, perform a forward pass of the tagger according to word (or sub-tree) representations, as well as constituent tags.
- Compute a *likelihood of the right sequence of BIOS-prefixed tags* (see below), given the scores of the tagger.
- Backward gradient through the tagger up to the word (or sub-tree) and tag representations.
- Backward gradient through the word-tag composer up to the word and tag representation.
- Update all model parameters (from compositional networks $C_i$, tagger network, and lookup tables) with a fixed learning rate.

Details about the training likelihood can be found in (Legrand & Collobert, 2014). The score $S([t]_1^N, [X]_1^N, \theta)$ of the true sequence of BIOS-prefixed tags $[t]_1^N$, given the input node sequence $[X]_1^N$ can be interpreted as a conditional probability by exponentiating this score (thus making it positive) and normalizing it with respect to all possible path scores. The log-probability of a sequence of tags $[t]_1^N$ for the input sequence of constituents $[X]_1^N$ is given by:

$$\log P([t]_1^N | [X]_1^N, \theta) = S([t]_1^N, [X]_1^N, \theta) \quad (1)$$
$$- \log \left[ \sum_{\forall [t']_1^N} \exp S([t']_1^N, [X]_1^N, \theta) \right].$$

The second term of this equation (which correspond to the normalisation term) can be computed in linear time thanks to a recursion similar to the Viterbi algorithm (see Rabiner, 1989). Similar training procedures have been proposed in the past for structured data (Denker & Burges, 1995; Bottou et al., 1997; Lafferty et al., 2001).

## 4 EXPERIMENTS

### 4.1 CORPUS

Experiments were conducted using the standard English Penn Treebank data set (Marcus et al., 1993). We adopted the classical setup, with sections 02-21 for train, section 22 for validation, and section 23 for test. The validation corpus was used to select our hyper-parameters and best models.

We pre-processed the data only with a subset of operations which are achieved in standard parsers: (1) functional labels, traces were removed, (2) the PRT label was replaced by ADVP (Magerman, 1995). (3) We tackled the unary chain issue - non-terminals with a single non-terminal child - by merging the nodes together and assigning as tag the concatenation of the merged node tags. This was done in order to avoid looping issues in the parsing algorithm (e.g. a node being repetitively tagged with two different tags in our iterative process) and ensure the convergence of the parsing process. Only concatenated labels which occurred at least 30 times (corresponding to the lowest number of occurrences of the less common original parsing tag) were kept, leading to 11 additional parsing tags. Added to the original 26 parsing tags, this resulted in 161 tags produced by our parser. At test time, the inverse operation is performed: concatenated tag nodes are simply expanded into their original form.

### 4.2 DETAILED SETUP

Our systems were trained using a stochastic gradient descent over the available training data until convergence on the validation set. Hyper-parameters were chosen according to the validation. Lookup-table sizes for the words and tags (part-of-speech and parsing) are 200 and 20, respectively. The window size for the tagger is $K = 7$ (3 neighbours from each side). The size of the tagger's hidden layer is $H = 500$. We used the word embeddings obtained from Lebret & Collobert (2014) to initialize the word lookup-table. These embeddings were then fine-tuned during the training process. We fixed the learning rate to $\lambda = 0.15$ during the stochastic gradient procedure. As suggested in Plaut & Hinton (1987), the learning rate was divided by the size of the input vector of each layer. The part-of-speech tags were obtained using the freely available software SENNA[4].

---
[4] http://ml.nec-labs.com/senna

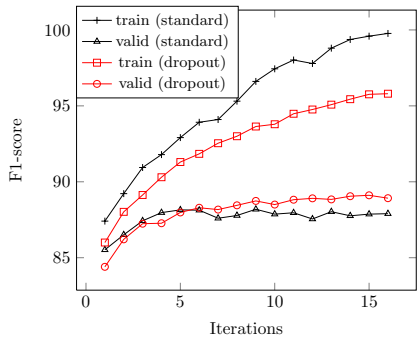 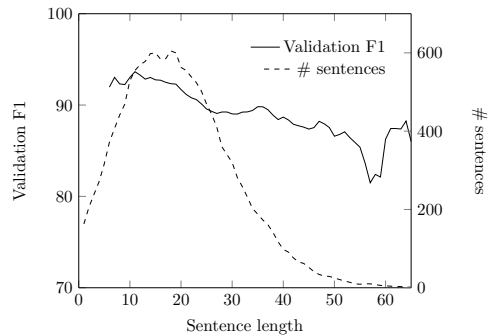

Figure 4: Train and validation F1-score, according to the number of training iterations, with and without the "dropout" procedure.

Figure 5: Validation F1 and number of sentences, according the the sentence length.

### 4.3 WORD EMBEDDING DROPOUT REGULARIZATION

We found that our system was easily subject to overfitting (training F1-score increasing while the validation curve was eventually decreasing as shown in Figure 4). As the capacity of our network mainly lies on the words and tag embeddings, we adopted a dropout regularization strategy (see Hinton et al., 2012) for the lookup tables. The key idea of the dropout regularization is to randomly drop units (along with their connections) from the neural network during training. This prevents units from co-adapting too much. In our case, during the training phase, a "dropout mask" is applied to the output of the lookup-tables: each element of the output is set to $0$ with a probability $0.25$. At test time, no patch is applied but the output is re-weighted, scaling it by $0.75$. We observed a good improvement in F1-score performance, as shown in Figure 4.

### 4.4 PERFORMANCE COMPARISON

F1 performance scores are reported in Table 1. Scores were obtained using the Evalb implementation[5]. We compared our system is compared with a range of different state-of-the-art parsers. In addition to the the four main generative parsers, we report the scores of well known re-ranking parsers (including the state-of-the-art from McClosky et al. (2006)) as well as for two major purely discriminative parsers. Detailed error analysis compared against a subset of these parsers is reported in Table 2, using the code provided by Kummerfeld et al. (2012). Performance with respect to sentence length is reported in Figure 5.

We included a voting procedure using several models trained starting from different random initializations. The voting procedure is achieved in the following way: at each iteration of the greedy parsing procedure, given the input sequence of constituents, (1) node representations are computed for each model by composing the sub-tree representations corresponding to the given model and using its own compositional network (2) each model computes tag scores using its own tagger network (3) tag scores are averaged (4) a coherent path of tag is predicted using the Viterbi algorithm.

Finally, we report a brief quantitative evaluation of our compositional representations in Table 3. Random phrases were picked in the WSJ corpus, and closest neighbors (according to the Euclidean distance) with other phrases of the corpus are reported.

---

[5] Available at http://nlp.cs.nyu.edu/evalb

Table 1: Performance comparison of different state-of-the-art parsers, in terms of Precision (P), Recall (R), and F1 score, for sentences of size $\leq 40$ words, and on the full WSJ test set. $V_x$ denotes a voting procedure with $x$ models. The reported time (in seconds) is the time to parse the full WSJ test corpus.

|  |  | < 40 | | | FULL | | | |
| --- | --- | --- | --- | --- | --- | --- | --- | --- |
|  | MODEL | (R) | (P) | F1 | (R) | (P) | F1 | TIME |
| GENERATIVE | MAGERMAN (1995) | 84.6 | 84.9 | 84.8 | | | | |
|  | COLLINS (1999) | 88.5 | 88.7 | 88.6 | 88.1 | 88.3 | 88.2 | 1247 |
|  | CHARNIAK (2000) | 90.1 | 90.1 | 90.1 | 89.6 | 89.5 | 89.6 | |
|  | PETROV & KLEIN (2007) | 90.7 | 90.5 | 90.6 | 90.2 | 98.9 | 90.1 | 307 |
| GENERATIVE WITH RE-RANKING | HENDERSON (2004) | | | | 89.8 | 90.4 | 90.1 | |
|  | CHARNIAK & JOHNSON (2005) | | | 92.0 | | | 91.1 | |
|  | MCCLOSKY ET AL (2006) | | | | | | **92.1** | |
|  | SOCHER ET AL (2013) | | | 91.1 | | | 90.4 | 390 |
| DISCRIMINATIVE | CARRERAS ET AL. (2008) | | | | 90.7 | 91.4 | 91.1 | |
|  | LEGRAND & COLLOBERT (2014) ($V_{10}$) | 90.0 | 90.1 | 90.1 | 89.6 | 89.7 | 89.6 | |
|  | LEGRAND & COLLOBERT (2014) + DROPOUT ($V_{10}$) | 90.6 | 90.1 | 90.4 | 90.2 | 89.7 | 89.9 | |
|  | THIS WORK | 88.8 | 89.1 | 89.0 | 88.2 | 88.6 | 88.4 | |
|  | THIS WORK + DROPOUT | 89.7 | 90.3 | 90 | 89.1 | 89.9 | 89.5 | 30 |
|  | THIS WORK + DROPOUT ($V_4$) | 90.5 | 90.8 | 90.7 | 90.1 | 90.4 | 90.3 | 120 |

Table 2: Detailed parser comparison. We report the average number of bracket errors per sentence for different error categories.

|  | PP ATTACH | CLAUSE ATTACH | DIFF LABEL | MOD ATTACH | NP ATTACH | CO-ORD | 1-WORD SPAN | UNARY | NP INT. | OTHER |
| --- | --- | --- | --- | --- | --- | --- | --- | --- | --- | --- |
| MCCLOSKY ET AL (2006) | 0.60 | 0.38 | 0.31 | 0.25 | 0.25 | 0.23 | 0.20 | 0.14 | 0.14 | 0.50 |
| SOCHER ET AL (2013) | 0.79 | 0.43 | 0.29 | 0.27 | 0.31 | 0.32 | 0.31 | 0.22 | 0.19 | 0.41 |
| LEGRAND ET AL (2014) | 0.74 | 0.45 | 0.27 | 0.25 | 0.34 | 0.38 | 0.24 | 0.22 | 0.20 | 0.57 |
| THIS WORK + DROPOUT | 0.78 | 0.44 | 0.29 | 0.27 | 0.36 | 0.42 | 0.24 | 0.21 | 0.20 | 0.60 |
| THIS WORK + DROPOUT ($V_4$) | 0.71 | 0.43 | 0.25 | 0.24 | 0.35 | 0.38 | 0.23 | 0.21 | 0.19 | 0.56 |

## 5 CONCLUSION

In this paper, we introduced a new parsing architecture which leverages RNN-based compositional representation of parsing sub-trees, both encoding the syntactic (tags) and semantic (words) information. The parsing procedure is tightly integrated with the composition operation, and allows us to reach performance of very well-known parsers while (1) adopting a greedy and fast procedure (2) avoid standard refined features such as headwords.

### ACKNOWLEDGMENTS

Part of this work was supported by NEC Laboratories America.

Table 3: Nearest neighbors (in terms of vector representation Euclidean distance) for several phrases in the WSJ corpus. For every node in the corpus, the sub-tree representations were computed. Then, for the selected phrases, we computed all Euclidean distances. We report below the 5 top closest other phrases in WSJ.

| |
|---|
| brendan barba , chairman of the moonachie , n.j. , maker of plastic film products |
| edmund edelman , chairman of the los angeles county board of supervisors |
| esther dyson , editor of release 0.0 , an industry newsletter that spots new developments |
| michael slater , editor of the microprocessor report , an industry newsletter |
| bruce miller , president of art funding corp. , an art lender |
| jeffrey nichols , president of apms canada , toronto precious metals advisers , |
| |
| eli lilly & co. , indianapolis , |
| john kinnard & co. , minneapolis , |
| procter & gamble co. , cincinnati , |
| anb investment management co. , chicago , |
| scimed life systems inc. , minneapolis , |
| rjr nabisco inc. 's french cracker subsidiary , belin , |
| |
| mr. engelken 's sister , martha , who was born two days before the home run , |
| the company 's president , n.j . nicholas , who will eventually be co-chief executive of time warner alongside mr. ross , |
| claudio 's sister , isabella , a novitiate in a convent , |
| her daughter , elizabeth , an attorney who is vice chairman , |
| his brother , parkhaji , whose head is swathed in a gorgeous crimson turban , |
| mrs. coleman 's husband , joseph , a physician , |
| |
| chairman and chief executive officer |
| president and chief executive officer |
| president and chief operating officer |
| chairman and chief executive |
| executive vice president and chief financial officer |
| executive vice president and chief operating officer |